%% file: main.tex
\documentclass[conference]{IEEEtran}

\newcommand{\cmmnt}[1]{}

\IEEEoverridecommandlockouts 
\usepackage{amsmath}

\usepackage{comment}
\usepackage{blindtext}
\usepackage{dirtytalk}
\usepackage{array}
\usepackage{authblk}
\usepackage{url}
\usepackage[noadjust]{cite}
\usepackage{graphics}
\usepackage{graphicx}
\usepackage{epsfig}
\usepackage{amssymb}
\usepackage{amsmath}
\usepackage{breqn}
\usepackage[ruled,vlined]{algorithm2e}

\usepackage{mathtools}
\usepackage{lscape}
\usepackage{pdflscape}
\usepackage{longtable}
\usepackage{makecell}
\usepackage{lettrine}
\usepackage{caption}
\usepackage{supertabular,booktabs}

\usepackage{amsmath}
\usepackage{graphicx}
\usepackage[colorlinks=true, allcolors=blue]{hyperref}
\usepackage{amssymb}
\usepackage[table,xcdraw]{xcolor}
\usepackage{tabularx}
\usepackage{multirow}

\usepackage{subfigure}
\usepackage{tikz}

\newcommand\copyrighttext{%
  \footnotesize \textcopyright \the\year{} IEEE. Personal use of this material is permitted.  Permission from IEEE must be obtained for all other uses, in any current or future media, including reprinting/republishing this material for advertising or promotional purposes, creating new collective works, for resale or redistribution to servers or lists, or reuse of any copyrighted component of this work in other works.}

\newcommand\copyrightnotice{%
\begin{tikzpicture}[remember picture,overlay]
\node[anchor=south,yshift=10pt] at (current page.south) {\fbox{\parbox{\dimexpr0.75\textwidth-\fboxsep-\fboxrule\relax}{\copyrighttext}}};
\end{tikzpicture}%
}

\newcolumntype{M}{>{$\displaystyle}c<{$}}
\newcommand\AddLabel[1]{\refstepcounter{equation}(\theequation)\label{#1}}
\newcolumntype{L}{>{\collectcell\AddLabel}r<{\endcollectcell}}% labeled

\title{\LARGE
Evaluating the Energy Efficiency of Few-Shot Learning for Object Detection in Industrial Settings}

\begin{document}

\author{Georgios Tsoumplekas\IEEEauthorrefmark{1}, Vladislav Li\IEEEauthorrefmark{2}, Ilias Siniosoglou\IEEEauthorrefmark{1}\IEEEauthorrefmark{3}, Vasileios Argyriou\IEEEauthorrefmark{2}, Sotirios K. Goudos\IEEEauthorrefmark{4},\\ Ioannis D. Moscholios\IEEEauthorrefmark{5}, Panagiotis Radoglou-Grammatikis\IEEEauthorrefmark{3}\IEEEauthorrefmark{6} and Panagiotis Sarigiannidis\IEEEauthorrefmark{1}\IEEEauthorrefmark{3}

\thanks{\IEEEauthorrefmark{1} G. Tsoumplekas, I. Siniosoglou and P. Sarigiannidis are with the R\&D Department, MetaMind Innovations P.C., Kozani, Greece - \texttt{E-Mail: \{gtsoumplekas, isiniosoglou, psarigiannidis\}@metamind.gr}}

\thanks{\IEEEauthorrefmark{2} V. Li and V. Argyriou are with the Department of Networks and Digital Media, Kingston University, Kingston upon Thames, United Kingdom - \texttt{E-Mail: \{v.li, vasileios.argyriou\}@kingston.ac.uk}}

\thanks{\IEEEauthorrefmark{3} I. Siniosoglou, P. R. Grammatikis and P. Sarigiannidis are with the Department of Electrical and Computer Engineering, University of Western Macedonia, Kozani, Greece - \texttt{E-Mail: \{isiniosoglou, pradoglou, psarigiannidis\}@uowm.gr}}

\thanks{\IEEEauthorrefmark{4} S. K. Goudos is with the Physics Department, Aristotle University of Thessaloniki, Thessaloniki, Greece - \texttt{E-Mail: sgoudo@physics.auth.gr}}

\thanks{\IEEEauthorrefmark{5} I. D. Moscholios is with the Department of Informatics and Telecommunications, University of Peloponnese, Tripoli, Greece - \texttt{E-Mail: idm@uop.gr}}

\thanks{\IEEEauthorrefmark{6} P. R. Grammatikis is with the Department of Research and Development, K3Y Ltd., Sofia, 1000, Bulgaria - \texttt{E-Mail: pradoglou@k3y.bg}}
}

\maketitle
\copyrightnotice
\thispagestyle{empty}
\pagestyle{empty}

%%%%%%%%%%%%%%%%%%%%%%%%%%%%%%%%%
%%%%%%%%%%%%%%%%%%%%%%%%%%%%%%%%%

\begin{abstract}
    \input{sections/00_abstract}
\end{abstract}


\section{Introduction}
\label{Introduction}
\input{sections/01_introduction}

\section{Related Work}
\label{relevant_work}
\input{sections/02_literature}

\section{Methodology}
\label{Methodology}
\input{sections/03_methodology}

\section{Experimental Results}
\label{Evaluation}
\input{sections/04_results}

\section{Conclusions}
\label{Conclusions}
\input{sections/05_conclusion}

\section*{Acknowledgement}
\input{sections/06_acknowledgement}

\bibliographystyle{IEEEtran}
\bibliography{bib_file}

\end{document}

%% file: sections/00_abstract.tex
In the ever-evolving era of Artificial Intelligence (AI), model performance has constituted a key metric driving innovation, leading to an exponential growth in model size and complexity. However, sustainability and energy efficiency have been critical requirements during deployment in contemporary industrial settings, necessitating the use of data-efficient approaches such as few-shot learning. In this paper, to alleviate the burden of lengthy model training and minimize energy consumption, a finetuning approach to adapt standard object detection models to downstream tasks is examined. Subsequently, a thorough case study and evaluation of the energy demands of the developed models, applied in object detection benchmark datasets from volatile industrial environments is presented. Specifically, different finetuning strategies as well as utilization of ancillary evaluation data during training are examined, and the trade-off between performance and efficiency is highlighted in this low-data regime. Finally, this paper introduces a novel way to quantify this trade-off through a customized Efficiency Factor metric.

%% file: sections/01_introduction.tex
Over the last few years, the adoption of Deep Learning (DL) techniques in a variety of applications, such as energy, construction, healthcare, security and others, has been rapidly accelerating in a wide spectrum of domains.  This growth is propelled by the need for high accuracy and automation in complex industrial environments, where precision and efficiency are paramount. These environments often present unique challenges that require robust and adaptive solutions, which DL techniques are particularly adept at providing. Furthermore, this growth is fuelled by the broader evolution of AI across various sectors, with the implementation and addition of new and enhanced computing and cognitive capabilities all contributing to more sophisticated and capable AI systems capable of tackling more difficult, general and complex tasks.

The global industrial ecosystem is increasingly incorporating AI capabilities for operational efficiency and innovation, particularly in areas like predictive maintenance, quality control, logistics, and supply chain optimization. DL models are crucial for identifying outliers, anomalies, and irregularities in environments, enabling decisive decision-making and proactive safety operations \cite{Sesis2022}. However, the innate nature of AI presents challenges, particularly regarding energy consumption. As complex models process larger datasets, the energy required to train and run these models increases, raising environmental, sustainability, and scalability concerns. Therefore, energy efficiency of AI systems is a critical area of focus, necessitating research to create more energy-efficient and cost-effective AI models without compromising performance and accuracy. Addressing this challenge is vital for the sustainable growth and integration of AI in industrial settings.

Few-shot learning \cite{tsoumplekas2024toward} (FSL) is a learning paradigm that enables models to learn from a limited amount of data and has emerged as a prominent solution to tackle the extensive resource demands of traditional and modern AI models. In the industrial ecosystem, where data can sometimes be scarce or expensive to acquire or are dependent to sensitivity and provenance limitations, FSL offers a viable pathway for efficient model training and optimization. This is especially important in the case of object detection \cite{Padilla2020} where data is not only subject to privacy regulations but also it is difficult to acquire large amounts of image data for specific scenarios. However, quantifying and enhancing the energy efficiency of FSL models remains a challenge.

This paper examines the energy efficiency of FSL algorithms in object detection, a crucial issue for industries seeking sustainable and cost-effective AI solutions. This work aims to jointly evaluate these algorithms' performance and energy consumption patterns, providing insights into the trade-off between these two objectives. In particular, this work systematically evaluates the energy efficiency of various finetuned YOLOv8 models, considering both detection performance and energy consumption. To evaluate the models, three benchmark image datasets from the volatile industrial environment are used, namely, i) a Personal Protective Equipment (PPE) detection dataset, ii) a Construction Safety detection dataset and iii) a Fire Detection dataset. The overall contributions of this paper can be summarized as follows:

\begin{itemize}
    \item Introduces a novel metric, Efficiency Factor, to quantify and correlate the energy consumption vs performance trade-off of FSL models.
    \item Presents a thorough evaluation of various finetuned models' performance against their energy efficiency.
    \item Creates an in-depth comparative study of the effect of different benchmark datasets on a widely used object detection model.
    \item Evaluates the efficacy of FSL as a prominent method to minimize training time and energy consumption in AI models.
\end{itemize}

The rest of this paper is organized as follows: the related work is discussed in Section \ref{relevant_work}, followed by an overview of the methodology in Section \ref{Methodology}. Section \ref{Evaluation} provides a comprehensive analysis of the available data while measuring the energy efficiency and performance of the models. Section \ref{Conclusions} offers concluding remarks.

%% file: sections/02_literature.tex
\subsection{Object Detection}
Object Detection has made great progress through the development of models such as YOLOv8, Mask RCNN, and Fast RCNN, each making distinct contributions to the area. YOLOv8, being a successor to YOLOv5, provides significant enhancements in terms of both accuracy and speed \cite{Talaat2023} However, the energy efficiency of YOLOv8 still remains a challenge, especially in edge computing situations. Mask RCNN is an extension of Faster RCNN \cite{Rane2023} that allows for pixel-level segmentation \cite{Sujatha2023}, which means it can accurately locate objects and perform instance segmentation. It performs exceptionally well in situations that demand precise detection, such as medical picture analysis. Nevertheless, the intricate structure of the system results in increased computational expenses, hence affecting its energy efficiency in settings with limited resources. Fast RCNN, an antecedent to Faster RCNN, integrates selective search with a deep CNN, thereby diminishing the duplication in feature computing \cite{Diwan2023}. Although it represented a notable advancement in the efficiency of object identification, its dependence on external region proposal methods hinders its speed and energy efficiency in comparison to more integrated models such as Faster RCNN.

\subsection{Few-Shot Object Detection}
While most FSL research has traditionally focused on image classification, in recent years there is an increasing interest in the development of novel few-shot object detection (FSOD) algorithms. One of the first approaches towards that direction has proposed a reweighting module that transforms the extracted feature representation and is jointly trained with a YOLO detection model in a two-step training procedure \cite{kang2019few}. A two-step training approach has also been explored in \cite{wang2020frustratingly} which demonstrates the effectiveness of finetuning in FSOD without the need for external modules. More recently, recasting the object detection problem as an image classification problem and learning new object classes in an adversarial manner has also been proposed \cite{li2021beyond}. Additionally, meta-learning has been leveraged to enable learning task-specific and task-agnostic model parameters in the context of FSOD \cite{wang2019meta}. Finally, FSOD has also been extended to novel settings via its combination with incremental learning \cite{yin2022sylph} and domain adaptation. %\cite{gao2023asyfod}

\subsection{AI Model Energy Efficiency}
The field of Green and Energy efficient AI is steadily evolving, with research currently focusing on mitigating the ecological consequences associated with training extensive ML models and aiming to quantize and subsequently tackle the energy needs of modern AI. In particular, the rise in model size and complexity has led to a large increase in energy usage and carbon emissions \cite{Henderson2020} and consequently, several methods have been suggested to reduce these impacts, such as calculating the carbon emissions caused by AI models and creating tools to assess the environmental impact of model training. Federated Learning (FL) has also been investigated as a means to mitigate energy usage \cite{qiu2023}. However, constraints in computational capacity and the requirement for inter-device communication present obstacles. Although Green AI is crucial, there remains a dearth of research on employing FSL as a viable alternative. While meta-learning has achieved excellent results in terms of performance in FSL, its large computational complexity renders it inhibiting for energy efficient applications. However, recently, transfer learning methods, such as fine-tuning, have emerged as viable alternatives in this context, due to their high performance and low computational cost.

%% file: sections/03_methodology.tex
\subsection{Model Architecture}
\label{architecture}

Given that our focus is towards models that adhere to the principles of Green AI\cite{schwartz2020green} and at the same time demonstrate strong generalization performance, YOLOv8, the latest version of the You-Only-Look-Once (YOLO) object detection models family, is used. Compared to its predecessor, YOLOv5, YOLOv8 introduces a highly efficient anchor-free object detection approach that leads to increased performance. 

Regarding its architecture, YOLOv8 consists of a backbone feature extractor, used to extract meaningful feature representations from the images, followed by the model's head network that produces the final predictions. As for the feature extractor, it is based on a modified version of CSPDarknet53\cite{bochkovskiy2020yolov4}. Its structure follows that of feature pyramid networks (FPNs) \cite{lin2017feature}, which enables the identification of objects of varying sizes and scales within an image by extracting features at multiple scales. On the other hand, the head network consists of a series of convolutional layers followed by three different detection modules whose inputs are features extracted from different levels of the FPN, allowing for multi-scaled object detection.

%%%%%%%%%%%%%%%%%%%%%%%%%%%%%%%%%
%%%%%%%%%%%%%%%%%%%%%%%%%%%%%%%%%
\subsection{Few-shot Object Detection }
\label{fsod}
The main objective of FSOD is training of models that are capable of quickly adapting to novel tasks given only a minimal number of training samples within each new task. In the context of FSOD, a standard approach is to consider two different sets of data with different object classes, the base classes and the novel classes. In this case, it is realistic to assume that the examined object detection model has been trained on a training set that consists of abundant data belonging to these base classes. More specifically, the training set can be denoted as:

\begin{equation}
    D^{train} = \{(x_i, y_i)\}_{i=1}^{|D^{train}|}
\end{equation}

\noindent where $x_i \in \mathbb{R}^M$ is the $i$-th training image, $y_i \in \{0, 1, ..., N_B-1\}$ is its label, and $N_B$ is the total number of base classes. Consequently, a trained model $f_{\theta}$ is produced.

Following training of the model on $D^{train}$, the next step is its adaptation to the novel classes. In general, these novel classes can be formulated as part of a few-shot task that consists of only a small of number of novel class images available during adaptation and an arbitrary number of novel class images used for evaluation of the model. Specifically, for a given task $\tau$ with $N_N$ novel classes, it can be split into a support set $\mathcal{S} = \{(x_i,y_i)\}_{i=1}^{|\mathcal{S}|}$ used for adaptation, and a query set $\mathcal{Q} = \{(x_j,y_j)\}_{j=1}^{|\mathcal{Q}|}$ used for model evaluation in this task. Adhering to the standard methodology of FSL, $N$-way $K$-shot tasks are examined, which consist of $N$ novel classes and there are $K$ training samples for each novel class in the task's support set (as a result $|\mathcal{S}| = NK$).

To adapt to task $\tau$ that contains the novel class samples, the trained model $f_{\theta}$ is further trained on $\tau$'s support set $\mathcal{S}$, producing an adapted model $f_{\theta'}$ which is then evaluated on $\tau$'s query set $\mathcal{Q}$. Finally, the model's overall performance is its mean performance across all tasks $\{\tau_i\}_{i=1}^{T}$, where $T$ is the total number of tasks, reported along with the corresponding standard deviation.

%%%%%%%%%%%%%%%%%%%%%%%%%%%%%%%%%
%%%%%%%%%%%%%%%%%%%%%%%%%%%%%%%%%
\subsection{Few-Shot Learning via Model Finetuning}
\label{model_finetuning}

\begin{figure}
 \begin{center}
  \includegraphics[width=0.48\textwidth]{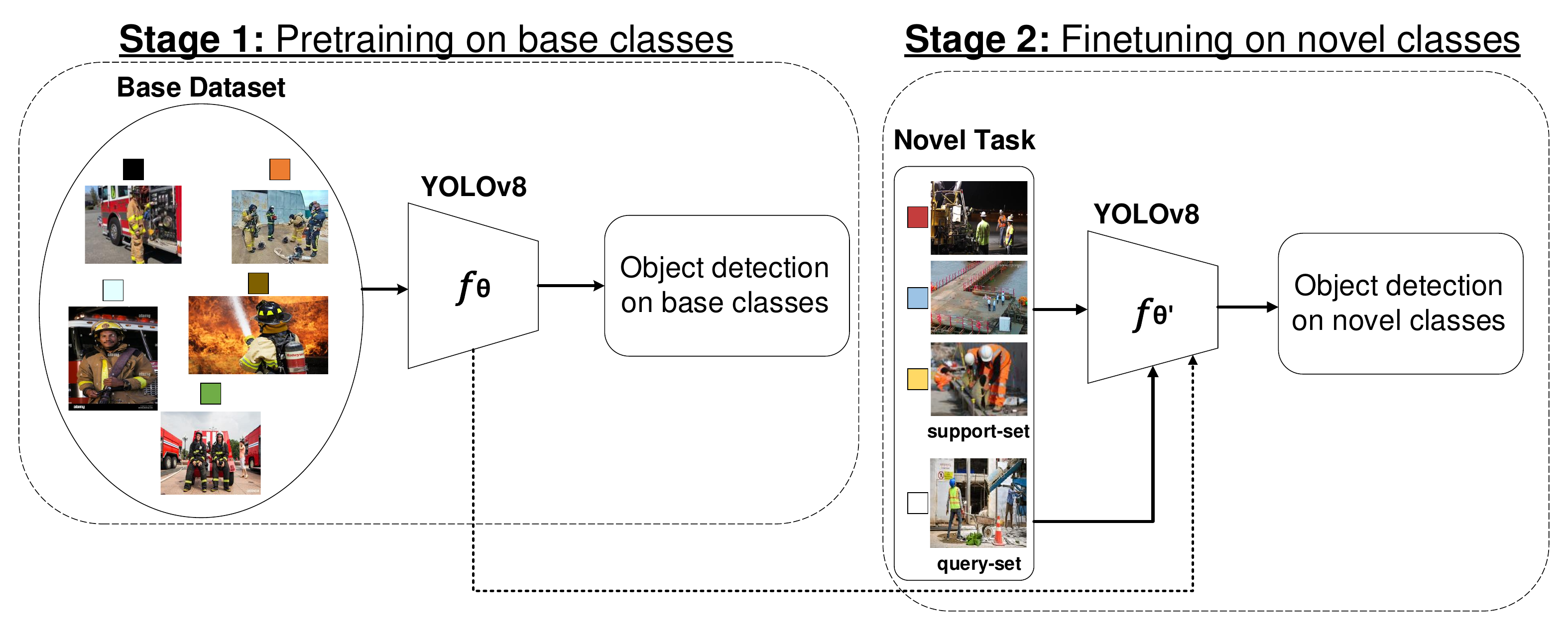}
 \end{center}
 \caption{Illustration of the two-step training procedure based on base class pretraining and novel class finetuning.}
 \label{two_step_finetuning}
\end{figure}

Based on the aforementioned standard methodology of FSOD, one question that typically arises in these settings is how to effectively and efficiently obtain $f_{\theta'}$ from $f_{\theta}$ using $\mathcal{S}$. One of the simplest approaches in this case, that has recently achieved competitive results compared to more complex solutions in various settings, is to simply finetune the trained model $f_{\theta}$ in $\mathcal{S}$. Overall, this finetuning approach can be seen as a two-step procedure. In the first step, the model is trained on $D^{train}$, emulating a form of model pretraining, and producing $f_{\theta}$. In the second step, the pretrained model is then finetuned on each task's support set $\mathcal{S}$, producing the final finetuned model $f_{\theta'}$. This two-step approach is also illustrated in Figure \ref{two_step_finetuning}.

A common consideration when finetuning is used to adapt a model in the context of FSL is deciding on a full vs partial finetuning approach. In general, both approaches have been successfully employed in this setting with one side arguing that adapting only the head of a model is enough to achieve competitive results, without imposing any additional computational burden, and the other side claiming that finetuning the backbone, too, is essential to avoid producing distorted feature representations due to the possible domain shift between base and novel classes.

To further examine these two different approaches both in terms of downstream task performance as well as computational efficiency during training, we employ three different model finetuning variations, leveraging YOLOv8's internal structure: (a) full finetuning of the whole model, including both backbone and head, (b) partial finetuning including the model's head only, and (c) partial finetuning including the model's detection modules only.

%% file: sections/04_results.tex
\subsection{Experimental Configuration}
\label{Testbed}

To allow for a fair comparison across the three aforementioned finetuning approaches, model pretraining is fixed in all cases and a pretrained version of YOLOv8 trained on the MSCOCO dataset is used, specifically YOLOv8n which consists of 3.2M parameters. In the case of full model finetuning all 3.2M parameters are adapted, while detection modules finetuning involves the adaptation of $\approx$750K parameters, and head finetuning involves the adaptation of $\approx$1.7M parameters. In our experiments, full finetuned models are denoted as $full$, models with finetuned heads are denoted with $head$, and models with finetuned detection modules are denoted with $det$.

Additionally, during model finetuning, object classes have only $K$ support set samples, with $K \in \{1,2,3,5,10,30\}$. The number of finetuning epochs is also adapted based on $K$. For $K=1$, the number of epochs is 10, for $K \in \{2,3,5\}$ it is 30, and finally for $K \in \{10, 30\}$ it is 200. The use of a validation set to measure model performance after each finetuning epoch is also examined (despite potential computational overhead) to facilitate selecting the best-performing model to be used during model testing. In our experiments, models using a validation set are denoted as $best$, while the rest are denoted as $last$. To ensure robust comparisons, each model is evaluated in three different downstream tasks, with the mean value and standard deviation reported for each metric across these tasks.

Finally, the optimized AI models are tested on an edge ecosystem, considering resource limitations. A resource-constrained edge device with a 12th Gen Intel i7 CPU, 16GB memory, integrated graphics, and Ubuntu 22.04 was used.

%%%%%%%%%%%%%%%%%%%%%%%%%%%%%%%%%
%%%%%%%%%%%%%%%%%%%%%%%%%%%%%%%%%
\subsection{Datasets}
For the following experiments, three different datasets related to object detection of objects commonly found in industrial settings were utilized to finetune the object detection models, in an effort to establish recognition and proactiveness in safety upkeeping and robust decision-making:

\begin{itemize}
    \item \textbf{Personal Protective Equipment Detection \cite{Sesis2022}:} Aims to help AI detectors locate and identify various PPE used by first responders for enhanced safety. 
    \item \textbf{Construction Safety Detection \cite{worker-safety_dataset2022}: }Trains AI models to identify PPE presence or absence in industrial/construction settings. 
    \item \textbf{Fire Detection:} Focuses on training AI to locate, recognize, and classify fires for proactive safety measures
\end{itemize}

\begin{figure}
 \begin{center}
  \includegraphics[width=0.32\textwidth]{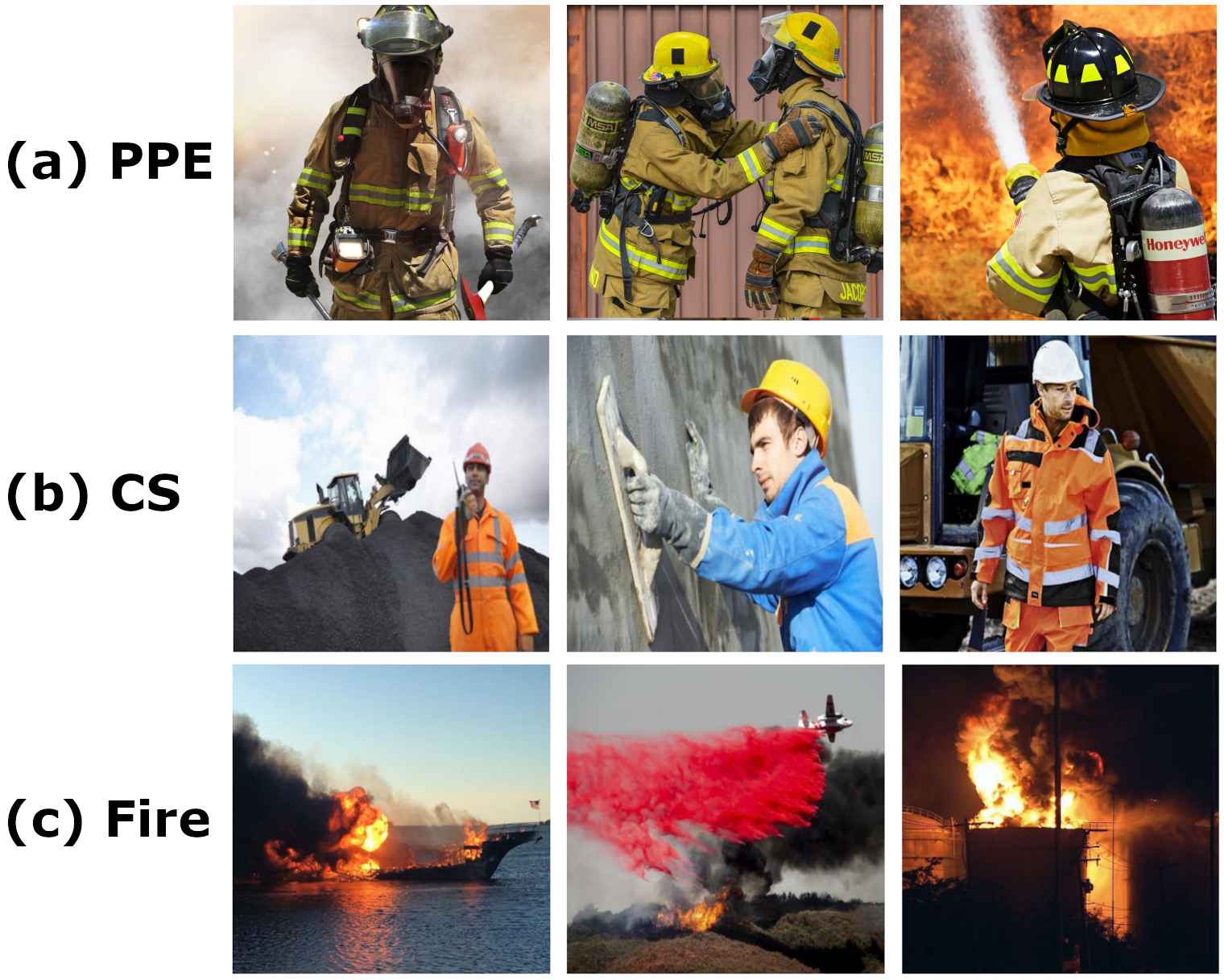}
 \end{center}
 \caption{Dataset Preview. (a) PPE, (b) Construction Safety, (c) Fire Detection }
 \label{Dataset_Preview}
\end{figure}

The abovementioned datasets were utilized to finetune the leveraged AI models of this work in the premise of evaluating the final models' performance vs energy efficiency for object detection. The attributes of these datasets are summarized in Table \ref{Dataset_Attributes}, while Figure \ref{Dataset_Preview} shows some examples of these datasets' images.

\begin{table*}[]
\centering
\caption{Main characteristics of the examined industrial object detection datasets.}
\label{Dataset_Attributes}
\resizebox{0.99\textwidth}{!}{%
\begin{tabular}{lllllll}
\textbf{Dataset Name}                   & \textbf{Total Samples} & \textbf{Classes}   & \textbf{}                                                              & \textbf{Train} & \textbf{Test} & \textbf{Validation} \\ \hline
PPE Detection Dataset                   & 342                    & Helmet, Gloves, Mask, Cloth & Localize and identify PPE for first responders                         & 280            & 31            & 31                  \\
Worker-Safety Computer   Vision Project & 3200                   & Helmets, Vests, other PPE   & Identify presence/absence of PPE in industrial settings & 2991           & 90            & 119                 \\
Fire Detection Dataset                  & 3677                   & Fire scenes                 & Locate, recognize, and classify fires                                  & 3527           & -             & 150                
\end{tabular}
}
\end{table*}

%%%%%%%%%%%%%%%%%%%%%%%%%%%%%%%%%
%%%%%%%%%%%%%%%%%%%%%%%%%%%%%%%%%
\subsection{Evaluation Metrics}

For the evaluation of the examined models, both performance and efficiency should be taken into consideration. In this setting, model performance refers to the typical evaluation metrics used to measure the generalization capabilities of a model on a given dataset and, consequently, can be quantified using Mean Average Precision (mAP):

\begin{equation}
    mAP = \frac{1}{N} \sum_{i=1}^N AP_i
\end{equation}

\noindent where $AP_i$ is the average precision for class $i$, and $N$ is the total number of classes in the dataset.

On the other hand, model efficiency refers to a measure of how energy-efficient is the examined model. Consequently, energy consumption, measured in watt-hours (Wh), during training of the model is used. Since all models used here leverage the same pretrained model, we only focus on the energy consumption during model finetuning. To compute energy consumption, CodeCarbon, a lightweight Python library suitable for emission and energy consumption tracking was used.

Finally, to account for both model performance and efficiency, a novel metric, Efficiency Factor ($EF$), is introduced that takes into consideration both mAP and energy consumption. Specifically, optimal models should demonstrate high performance (high mAP) as well as high efficiency (low energy consumption). As a result, $EF$ can be formulated as:

\begin{equation}
    EF = \frac{mAP}{1+EC}
\end{equation}

\noindent where $mAP \in [0, 100]$ is in its percentage form, and $EC \in (0, +\infty)$ is the model's energy consumption measured in $Wh$. A constant value is also included in the denominator so that $EF$ is bounded in $[0, 100)$, with higher $EF$ values given to models that achieve both high mAP values during testing and low energy consumption during training.

%%%%%%%%%%%%%%%%%%%%%%%%%%%%%%%%%
%%%%%%%%%%%%%%%%%%%%%%%%%%%%%%%%%
\subsection{Experimental Results}
\label{Experiment_Results}

\begin{table}[]
\centering
\caption{Few-shot detection performance (mean mAP and standard deviation) on each of the three examined datasets.}
\label{mAP}
\resizebox{0.99\columnwidth}{!}{%
\begin{tabular}{llcccccc}
\hline
\multicolumn{1}{c}{\multirow{2}{*}{\textbf{Dataset}}} & \multicolumn{1}{c}{\multirow{2}{*}{\textbf{Model}}} & \multicolumn{6}{c}{\textbf{Shots}}                                      \\ \cline{3-8}
        & & \textbf{1} & \textbf{2} & \textbf{3} & {\textbf{5}} & \multicolumn{1}{c}{\textbf{10}} & \textbf{30} \\ \hline
        \multicolumn{1}{c}{\multirow{6}{*}{\textbf{PPE}}}  & 
        y8-det-last & 9.68±1.43 & 6.79±2.35 & 10.49±1.94 & 10.93±1.10 & 16.50±1.89 & 19.11±1.12 \\
        & y8-det-best & 9.71±1.39 & \textbf{10.22±1.51} & 12.47±1.41 & \textbf{13.10±0.64} & 16.21±0.72 & 21.29±0.94 \\
        & y8-head-last & 9.97±2.04 & 6.84±2.32 & 10.95±1.66 & 10.91±1.88 & 19.63±4.09 & 29.88±1.49 \\
        & y8-head-best & \textbf{10.01±1.99} & 9.66±1.14 & 11.53±1.88 & 12.46±0.47 & \textbf{20.92±2.05} & 29.55±1.12 \\
        & y8-full-last & 9.69±1.07 & 8.84±2.20 & 12.76±3.06 & 10.07±3.78 & 13.09±2.07 & 29.87±1.52 \\
        & y8-full-best & 9.69±1.07 & 9.91±2.26 & \textbf{12.79±2.71} & 12.59±1.60 & 14.68±2.58 & \textbf{31.96±1.59} \\
        \hline
        \multicolumn{1}{c}{\multirow{6}{*}{\textbf{Fire}}}  & 
        y8-det-last & 1.63±0.24 & 1.99±0.49 & 1.75±0.15 & 1.74±0.42 & 1.54±0.49 & 1.63±0.53 \\
        & y8-det-best & 1.81±0.03 & 1.94±0.07 & 1.93±0.17 & 2.09±0.35 & 2.42±0.26 & 3.15±0.42 \\
        & y8-head-last & 1.69±0.20 & 2.04±0.27 & 1.95±0.27 & 1.73±0.36 & 0.79±0.34 & 1.12±0.76 \\
        & y8-head-best & \textbf{1.82±0.05} & \textbf{2.11±0.26} & 1.77±0.07 & \textbf{2.18±0.32} & \textbf{2.82±0.55} & \textbf{3.62±0.69} \\
        & y8-full-last & 1.61±0.15 & 1.39±0.27 & 1.75±0.98 & 0.86±0.38 & 0.73±0.40 & 0.84±0.69 \\
        & y8-full-best & \textbf{1.82±0.06} & 1.95±0.21 & \textbf{2.51±0.86} & 1.56±0.40 & 2.61±0.34 & 2.52±1.69 \\ \hline
        \multicolumn{1}{c}{\multirow{6}{*}{\textbf{CS}}}  & 
        y8-det-last & 8.54±0.98 & 10.30±3.41 & 10.88±2.64 & 12.79±2.70 & 19.48±3.03 & 31.64±0.74 \\
        & y8-det-best & 8.62±0.85 & 11.44±2.91 & 12.67±1.65 & 13.81±1.60 & 17.17±2.46 & 32.02±1.95 \\
        & y8-head-last & 8.77±1.43 & 11.92±0.76 & 9.57±3.16 & 11.31±1.94 & \textbf{26.16±0.42} & 36.28±1.74 \\
        & y8-head-best & \textbf{8.96±1.17} & \textbf{13.41±1.13} & 12.59±2.02 & 12.86±1.63 & 23.49±4.12 & \textbf{36.76±1.97} \\
        & y8-full-last & 8.81±1.37 & 11.70±0.80 & 10.15±2.77 & 13.00±2.57 & 24.76±5.39 & 34.54±3.91 \\
        & y8-full-best & 8.92±1.23 & 13.30±1.22 & \textbf{13.09±1.89} & \textbf{13.89±1.66} & 24.67±5.21 & 34.19±4.06 \\ \hline
\end{tabular}
}
\end{table}

\begin{figure}[htp]
  \centering
  \subfigure[CS Dataset]{
    \includegraphics[scale=0.45]{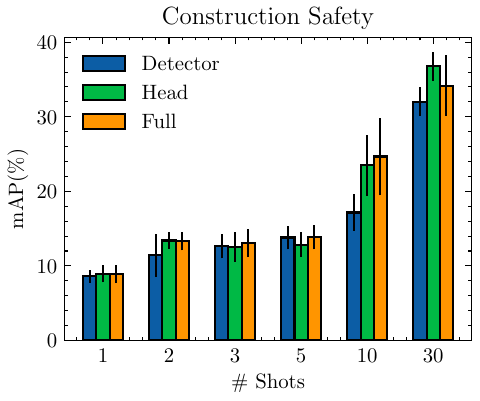}
    }
  \subfigure[Fire Dataset]{\includegraphics[scale=0.45]{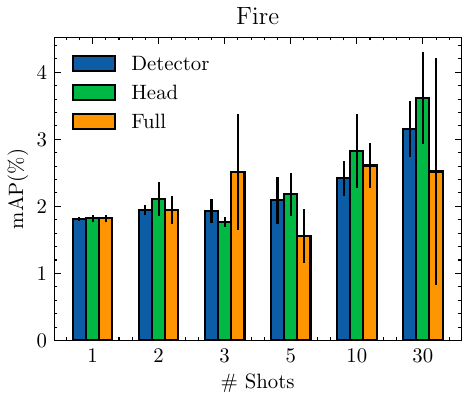}}
  \subfigure[PPE Dataset]{\includegraphics[scale=0.45]{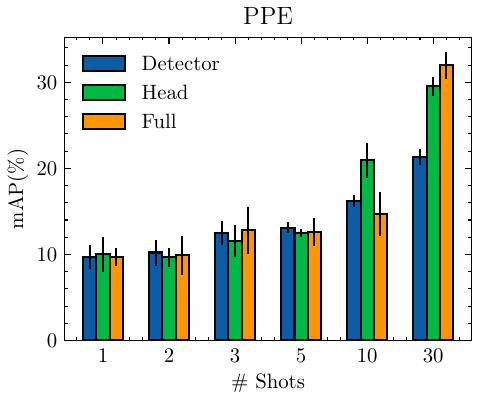}}
  \caption{Mean mAP and standard deviation of $best$ models for varying finetuning strategies and numbers of shots.}
  \label{map_barplot}
\end{figure}

\textbf{Model Performance. } Table \ref{mAP} shows the performance of all six different finetuning combinations in terms of mAP for the three examined datasets. It is evident that in almost all cases, using a validation set during finetuning leads to consistently better-performing models, outlining the importance of carefully selecting the number of finetuning epochs. At the same time, increasing the number of shots also leads to increased model performance, which is reasonable considering that there are more available training data in the task's support set. Finally, model performance seems to be less sensitive in the selection of the finetuning strategy with all three strategies leading to comparable results for the same number of shots. This is also illustrated in Figure \ref{map_barplot}, where the performance of models using evaluation sets during finetuning can be seen for different finetuning strategies and different numbers of shots.

\textbf{Model Efficiency. } Table \ref{energy_consumption} shows the energy consumption of the evaluated models during finetuning, clearly demonstrating that finetuning only the detection modules without using a validation set consistently outperforms the rest of the models in all three datasets and for each number of shots. Overall, models that do not use a validation set demonstrate better energy efficiency compared to the corresponding ones that use, which is logical given that finetuning becomes less computationally expensive without the use of the validation set in each finetuning epoch. Additionally, reducing the number of model parameters that are finetuned also leads to reduced energy consumption since the finetuning procedure becomes less computationally heavy.

\begin{table}[]
\centering
\caption{Energy consumption (mean Wh and standard deviation) on each of the three examined datasets.}
\label{energy_consumption}
\resizebox{0.99\columnwidth}{!}{%
\begin{tabular}{llcccccc}
\hline
\multicolumn{1}{c}{\multirow{2}{*}{\textbf{Dataset}}} & \multicolumn{1}{c}{\multirow{2}{*}{\textbf{Model}}} & \multicolumn{6}{c}{\textbf{Shots}}                                      \\ \cline{3-8}
        & & \textbf{1} & \textbf{2} & \textbf{3} & {\textbf{5}} & \multicolumn{1}{c}{\textbf{10}} & \textbf{30} \\ \hline
        \multicolumn{1}{c}{\multirow{6}{*}{\textbf{PPE}}}  & 
        y8-det-last & \textbf{0.271±0.003} & \textbf{0.639±0.036} & \textbf{0.843±0.007} & \textbf{1.194±0.060} & \textbf{12.435±0.268} & \textbf{30.010±0.658} \\
        & y8-det-best & 0.774±0.020 & 2.400±0.016 & 2.607±0.007 & 2.916±0.053 & 24.738±0.442 & 42.395±1.153 \\
        & y8-head-last & 0.282±0.002 & 0.720±0.032 & 0.994±0.001 & 1.464±0.066 & 16.821±0.359 & 41.395±0.715 \\
        & y8-head-best & 0.779±0.015 & 2.280±0.045 & 2.668±0.017 & 3.070±0.047 & 27.647±0.185 & 52.831±0.691 \\
        & y8-full-last & 0.360±0.004 & 1.095±0.068 & 1.517±0.014 & 2.196±0.096 & 25.169±0.290 & 63.614±0.919 \\
        & y8-full-best & 0.852±0.003 & 2.762±0.020 & 3.190±0.017 & 3.866±0.077 & 36.599±0.224 & 74.630±1.195 \\
        \hline
        \multicolumn{1}{c}{\multirow{6}{*}{\textbf{Fire}}}  & 
        y8-det-last &\textbf{0.487±0.010} & \textbf{0.601±0.005} & \textbf{0.660±0.007} & \textbf{0.734±0.004} & \textbf{3.772±0.040} & \textbf{10.294±0.062} \\
        & y8-det-best & 2.161±0.041 & 6.341±0.015 & 6.448±0.036 & 6.573±0.011 & 43.335±0.470 & 48.161±0.434 \\
        & y8-head-last & 0.545±0.009 & 0.741±0.012 & 0.783±0.009 & 0.897±0.013 & 5.709±0.021 & 13.889±0.110 \\
        & y8-head-best & 2.316±0.069 & 6.734±0.020 & 6.775±0.032 & 6.809±0.013 & 46.277±0.760 & 50.452±1.361 \\
        & y8-full-last & 0.532±0.012 & 0.782±0.014 & 0.852±0.009 & 1.018±0.013 & 7.235±0.018 & 20.388±0.504 \\
        & y8-full-best & 2.273±0.014 & 6.421±0.044 & 6.472±0.039 & 6.576±0.113 & 43.477±0.259 & 54.348±1.413 \\ \hline
        \multicolumn{1}{c}{\multirow{6}{*}{\textbf{CS}}}  & 
        y8-det-last & \textbf{0.443±0.007} & \textbf{0.857±0.011} & \textbf{1.095±0.025} & \textbf{1.677±0.028} & \textbf{15.681±0.088} & \textbf{43.325±0.673} \\
        & y8-det-best & 1.855±0.077 & 5.588±0.031 & 5.620±0.053 & 6.111±0.160 & 44.621±0.703 & 70.761±1.686 \\
        & y8-head-last & 0.495±0.006 & 1.032±0.008 & 1.365±0.041 & 2.033±0.013 & 20.858±0.384 & 57.974±0.673 \\
        & y8-head-best & 1.801±0.045 & 5.506±0.025 & 5.778±0.037 & 6.300±0.102 & 49.440±0.402 & 85.889±0.524 \\
        & y8-full-last & 0.616±0.009 & 1.468±0.018 & 1.913±0.067 & 3.093±0.031 & 33.436±0.776 & 88.492±0.635 \\
        & y8-full-best & 1.986±0.097 & 5.977±0.045 & 6.222±0.040 & 7.263±0.079 & 62.464±1.427 & 119.86±2.998 \\ \hline
\end{tabular}
}
\end{table}

\begin{figure}[htp]
  \centering
  \subfigure[CS Dataset]{\includegraphics[scale=0.45]{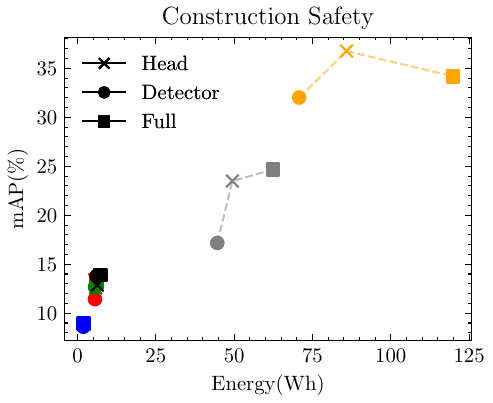}}
  \subfigure[Fire Dataset]{\includegraphics[scale=0.45]{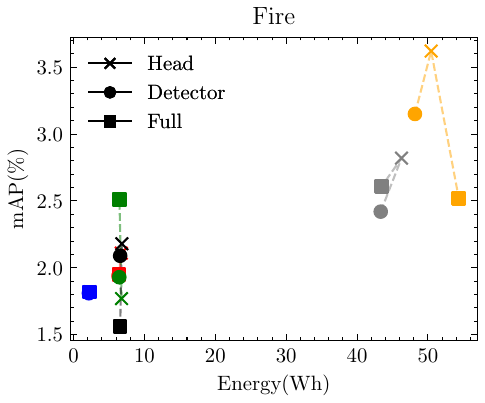}}
  \subfigure[PPE Dataset]{\includegraphics[scale=0.45]{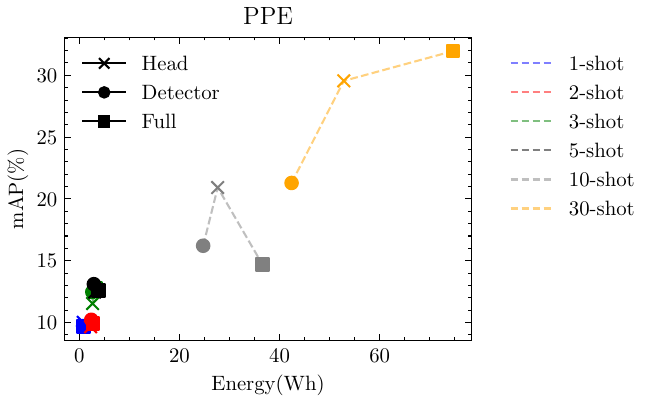}}
  \caption{mAP with respect to energy consumption for different finetuning strategies and numbers of shots.}
  \label{best_energy_map}
\end{figure}

\textbf{Effect of finetuning. } Figure \ref{best_energy_map} illustrates model performance with respect to the energy consumed during training for each of the examined datasets. Overall, optimal models in terms of performance and energy efficiency should achieve high mAP combined with low energy consumption. As a result, models that lie on the upper left area of each scatter plot are considered best. However, there is a clear trade-off between performance and efficiency since better-performing models consume more energy during finetuning. Finally, it is also worth noticing that finetuning the model's head seems to constitute a good compromise between these two conflicting objectives, especially for an increased number of shots.

\begin{figure}[h]
  \centering
  \subfigure[Detection modules finetuning]{\includegraphics[scale=0.45]{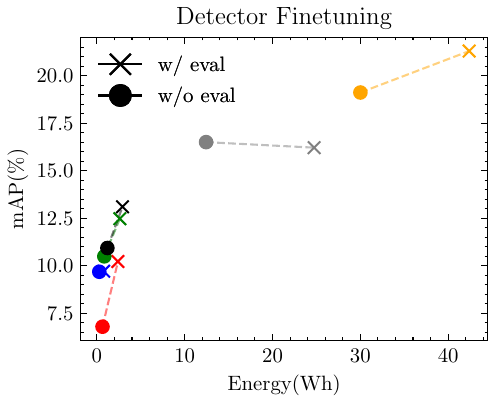}}
  \subfigure[Head finetuning]{\includegraphics[scale=0.45]{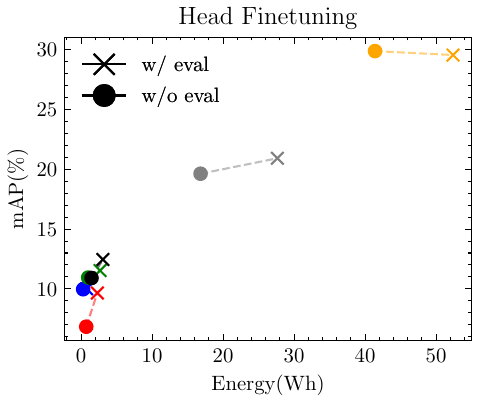}}
  \subfigure[Full finetuning]{\includegraphics[scale=0.45]{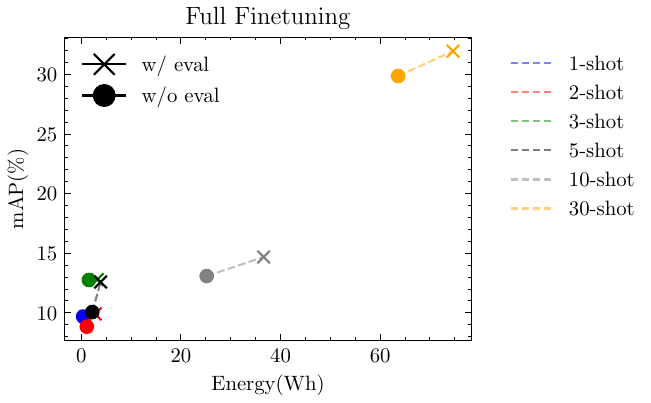}}
  \caption{Model performance with respect to energy consumption in the PPE dataset.}
  \label{ppe_energy_map}
\end{figure}

\textbf{Effect of validation set. } To illustrate how model performance with respect to energy consumption is affected by the use of a validation set during finetuning the related results from the PPE dataset are displayed in Figure \ref{ppe_energy_map}. Similar to Figure \ref{best_energy_map}, there seems to be a clear trade-off between model performance and energy consumption, underlining the challenges in developing FSOD models that are both sustainable and effective. Additionally, in terms of mAP, the use of a validation set becomes less important as the number of shots increases. However, its use leads to a significant increase in energy consumption, rendering it less useful in these scenarios.

\begin{figure}[h]
  \centering
  \subfigure[CS Dataset]{\includegraphics[scale=0.45]{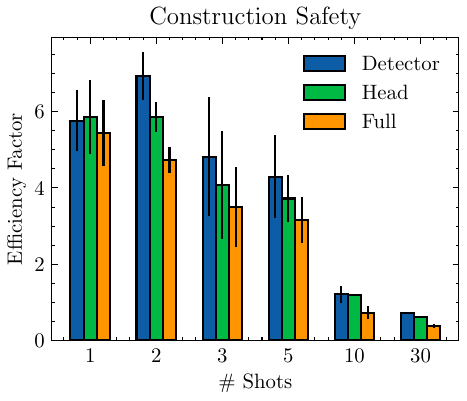}}
  \subfigure[Fire Dataset]{\includegraphics[scale=0.45]{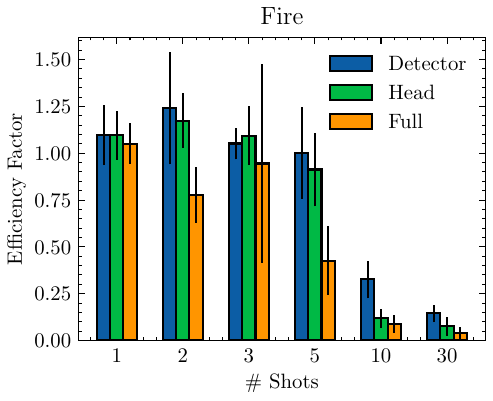}}
  \subfigure[PPE Fataset]{\includegraphics[scale=0.45]{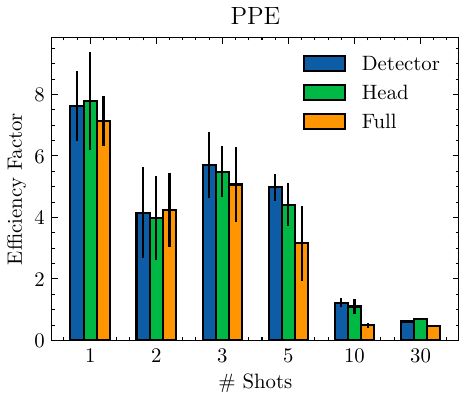}}
  \caption{Mean $EF$ and standard deviation of $last$ models for varying finetuning strategies and numbers of shots.}
  \label{ef_last}
\end{figure}

\textbf{EF as a consolidated performance-efficiency measure.} While performance and training efficiency seem to be two conflicting objectives in the context of FSOD, it is important to be able to compare models in a unified way, taking into consideration both of these desired properties. Table \ref{efficiency_factor} shows the $EF$ values of the examined models in all three datasets for a varying number of shots. It is manifest that not using a validation set during finetuning leads to increased $EF$ values due to the reduced energy consumption of these models compared to the corresponding ones that use validation sets. Additionally, while adapting less parameters during finetuning shrinks energy consumption, it does not necessarily lead to increased $EF$ values, since it might also be accompanied by a loss in performance. However, the number of shots directly affects $EF$, with a smaller number of shots leading to increased $EF$ values. This can be attributed to the disproportionate increase in energy consumption as the number of shots increases because of the corresponding escalation of the finetuning epochs. The aforementioned observations are also illustrated in Figure \ref{ef_last}, where the $EF$ values of models finetuned without using a validation set are displayed.

\begin{table}[]
\centering
\caption{Model performance per units of energy efficiency (mean $EF$ and standard deviation) on each of the three examined datasets.}
\label{efficiency_factor}
\resizebox{0.99\columnwidth}{!}{%
\begin{tabular}{llcccccc}
\hline
\multicolumn{1}{c}{\multirow{2}{*}{\textbf{Dataset}}} & \multicolumn{1}{c}{\multirow{2}{*}{\textbf{Model}}} & \multicolumn{6}{c}{\textbf{Shots}}                                      \\ \cline{3-8}
        & & \textbf{1} & \textbf{2} & \textbf{3} & {\textbf{5}} & \multicolumn{1}{c}{\textbf{10}} & \textbf{30} \\ \hline
        \multicolumn{1}{c}{\multirow{6}{*}{\textbf{PPE}}}  & 
        y8-det-last & 7.618±1.141 & 4.151±1.475 & \textbf{5.696±1.077} & \textbf{4.977±0.439} & \textbf{1.230±0.150} & 0.617±0.044 \\
        & y8-det-best & 5.484±0.828 & 3.008±0.453 & 3.456±0.385 & 3.347±0.208 & 0.630±0.026 & 0.491±0.032 \\
        & y8-head-last & \textbf{7.784±1.605} & 3.982±1.363 & 5.488±0.832 & 4.417±0.698 & 1.106±0.248 & \textbf{0.705±0.033} \\
        & y8-head-best & 5.635±1.154 & 2.948±0.374 & 3.146±0.519 & 3.060±0.079 & 0.730±0.071 & 0.549±0.027 \\
        & y8-full-last & 7.132±0.808 & \textbf{4.255±1.204} & 5.071±1.226 & 3.160±1.228 & 0.500±0.079 & 0.462±0.024 \\
        & y8-full-best & 5.235±0.575 & 2.637±0.605 & 3.052±0.647 & 2.590±0.348 & 0.390±0.067 & 0.423±0.027 \\
        \hline
        \multicolumn{1}{c}{\multirow{6}{*}{\textbf{Fire}}}  & 
        y8-det-last & 1.096±0.159 & \textbf{1.241±0.301} & 1.052±0.085 & \textbf{1.002±0.244} & \textbf{0.326±0.101} & \textbf{0.145±0.047} \\
        & y8-det-best & 0.573±0.003 & 0.265±0.010 & 0.260±0.023 & 0.276±0.047 & 0.055±0.006 & 0.064±0.008 \\
        & y8-head-last & \textbf{1.097±0.131} & 1.173±0.146 & \textbf{1.094±0.157} & 0.913±0.197 & 0.119±0.050 & 0.075±0.050 \\
        & y8-head-best & 0.549±0.007 & 0.272±0.033 & 0.227±0.010 & 0.280±0.041 & 0.060±0.013 & 0.070±0.013 \\
        & y8-full-last & 1.051±0.108 & 0.777±0.147 & 0.945±0.531 & 0.426±0.183 & 0.089±0.049 & 0.039±0.032 \\
        & y8-full-best & 0.556±0.019 & 0.263±0.028 & 0.336±0.114 & 0.206±0.054 & 0.059±0.008 & 0.046±0.030 \\ \hline
        \multicolumn{1}{c}{\multirow{6}{*}{\textbf{CS}}}  & 
        y8-det-last & 5.762±0.801 & \textbf{6.936±0.637} & \textbf{4.819±1.566} & \textbf{4.295±1.090} & \textbf{1.211±0.215} & \textbf{0.726±0.015} \\
        & y8-det-best & 3.009±0.272 & 2.065±0.234 & 1.866±0.287 & 1.836±0.289 & 0.397±0.063 & 0.452±0.026 \\
        & y8-head-last & \textbf{5.868±0.975} & 5.866±0.398 & 4.067±1.415 & 3.724±0.620 & 1.197±0.034 & 0.615±0.023 \\
        & y8-head-best & 3.195±0.367 & 2.061±0.176 & 1.855±0.287 & 1.759±0.205 & 0.466±0.085 & 0.423±0.023 \\
        & y8-full-last & 5.452±0.864 & 4.742±0.338 & 3.508±1.046 & 3.171±0.602 & 0.723±0.171 & 0.386±0.046 \\
        & y8-full-best & 2.978±0.322 & 1.907±0.185 & 1.813±0.273 & 1.679±0.184 & 0.391±0.089 & 0.282±0.026 \\ \hline
\end{tabular}
}
\end{table}

%% file: sections/05_conclusion.tex
While modern machine and deep learning approaches have led to tremendous strides towards operational efficiency, procedure optimization, and safety enhancement in various industrial settings and applications, their adaptation to realistic volatile industrial environments with scarce available data has been limited. At the same time, recent research developments have mainly focused on performance increase overlooking energy efficiency which is also a critical factor in these scenarios. In this paper, we examine how few-shot learning can be leveraged in the context of object detection in industrial settings to produce models that demonstrate both high performance and energy efficiency. Using a finetuning-based approach, an object detection model is adapted to downstream detection tasks with limited samples. An empirical study based on three different industrial datasets is conducted, demonstrating the trade-off between model performance and energy efficiency, while also examining how these variables are affected by different finetuning strategies. Finally, a novel metric, Efficiency Factor, is introduced to help quantify the interaction of model performance and efficiency in a consolidated way.

%% file: sections/06_acknowledgement.tex
This project has received funding from the European Union’s Horizon Europe research and innovation programme under grant agreement No. 101070181 (TALON).